# A Novel Comprehensive Approach for Estimating Concept Semantic Similarity in WordNet

Xiao-gang Zhang[1], Shou-qian Sun[2], Ke-jun Zhang[3*]

([1] *College of Computer, Zhejiang University, Hangzhou 310027, China /College of Information Engineering, Tarim University, Alar 843300, China*)

([2] *College of Computer, Zhejiang University, Hangzhou 310027, China*)

([3] *College of Computer, Zhejiang University, Hangzhou 310027, China*)

**Abstract:** Computation of semantic similarity between concepts is an important foundation for many research works. This paper focuses on IC (information content) computing methods and IC measures, which estimate the semantic similarities between concepts by exploiting the topological parameters of the taxonomy. Based on analyzing representative IC computing methods and typical semantic similarity measures, we propose a new hybrid IC computing method. Through adopting the parameter *dhyp* and *lch*, we utilize the new IC computing method and propose a novel comprehensive measure of semantic similarity between concepts. An experiment based on WordNet "is a" taxonomy has been designed to test representative measures and our measure on benchmark dataset R&G, and the results show that our measure can obviously improve the similarity accuracy. We evaluate the proposed approach by comparing the correlation coefficients between five measures (the proposed approach, four other similarity methods) and the artificial data. The results show that our proposal outperforms the previous measures.

## 1 Instruction

Computation of semantic similarity has already become the precondition for some research in various fields, including natural language processing, artificial intelligence, knowledge management and information retrieval [1]. The computation of concept semantic similarities is the fundamental for estimating textual semantic similarities because the concept is the smallest unit of semantic computing and the basis of information resource matching [2]. Utilizing the uniqueness of ontology concept and linguistic independence, polysemy and synonym of the concept can be effectively eliminated [3].

WordNet and the Wikipedia Category Graph (WCG) are both reference ontology in the computation of concept semantic similarity. WordNet is a universal semantic lexicon and common ontology, which developed by the Cognitive Science Laboratory of Princeton University [4]. Because of its versatility and rigorous semantic organization, WordNet has been implemented as the underlying reference ontology in various tasks of natural language processing, such as machine translation, word discrimination, keyword retrieval, text mapping, information extracting and so on. The WCG is the other resources in some works, including in works of Hadj Taieb et al. [5, 6] and Zesch [7]. The WCG is different from WordNet because it is proposed by volunteers, and the categories of WCG do not include specifying the type in semantic relations. In the paper, we adopt WordNet as the reference ontology [1].

In this paper, we propose a new comprehensive approach based on taxonomical parameters which are extracted from WordNet "is a" taxonomy, the taxonomical parameters including subsumer and implicating hyponyms, the depth ratio and the deepest common hypernym between the two concepts concerned by the semantic similarity task [1]. We utilize the proposed approach for computing the semantic similarities between words.

The rest of the paper is organized as follows. In part 2, we will analyze some related works, including typical semantic similarity measures and representative IC computing models. In part 3, we will improve existed IC model, propose our semantic similarity approach and apply this measure to benchmark dataset. In part 4, we will design an evaluation metrics to evaluate semantic similarity measures. In part 5, we will make an experiment to estimate the semantic similarities and compare the correlation coefficient of several similarity measures and artificial discrimination. In part 6, we will discuss the results of experiment and the contrast of the correlation. In part 7, we will take a conclusion to this paper and make a plan for future works.

## 2 Related works

Nowadays, some scholars at home and abroad have carried on extensive exploration and research on the concept similarity computation, and proposed many semantic similarity methods. Representative methods included IC-based measures, distance-based measures, feature-based measures and hybrid measures. The measure of IC-based computed concept similarity by examining the information content contained in the word pairs [8]. The measure of distance-based calculated concept similarity by the semantic distance (the number of edges linking two concepts) between words, and then transformed the distance into similarity value [9]. The measure of feature-based estimated the semantic similarities between words according to the structural feature of taxonomy, which included nodes and edges [10]. Hybrid measures computed the similarities between words by merging the advantages of other measures conceived [11].

In this paper, we focus on IC-based semantic similarity measures [12–15], which include two parts: computing IC method and IC-based similarity measures [10].

### 2.1 Estimating the IC of a concept

Computing IC is the key of computing IC-based similarity. Method of computing IC is usually divided into two categories according to the different calculating objects, one based on statistical information and the other based on ontology intrinsic structure.

### 2.1.1 The IC model based on statistical information

This method computed the value of IC by counting the probability of a concept in a given corpus. In this kind of method, the most typical model is the model proposed by Resnik. He put forward that the frequency of concept could be estimated by the term frequency appearing in Brown Corpus [15]. Resnik proposed the equation as follows [12]:

$$IC(c) = -log(p(c)) \quad (1)$$

Here, $c$ is a concept node, and $p(c)$ is the probability that $c$ appears in a given corpus. From equation (1), we can see that the more frequency of concept appeared, the less message of the concept transferred. Each term appeared in the corpus was counted as an occurrence rate of concept which included the term in ontology taxonomy. Then, $Freq(c)$ is computed as follows:

$$Freq(c) = \sum_{\omega \in Word(c)} Count(\omega) \quad (2)$$

Here, $Word(c)$ is a set of words subsumed by $c$, and $Count(\omega)$ represents the frequency of the word $\omega$ appeared in the corpus. Then, $p(c)$ is computed as follows [12]:

$$p(c) = \frac{Freq(c)}{N} \quad (3)$$

Wherein, $N$ is the total number of terms appeared.

In general, the advantages of this method are high efficiency and suitable for large-scale data processing. The shortcomings are subject to external interference and inaccurate value.

**2.1.2 The IC model based ontology intrinsic structure**

Compared with the model based on statistical information, this kind of method calculated the value of IC by the ontology intrinsic structure regardless of the external factor, but this kind of method asked the ontology taxonomy has been organized with a meaningful way.

Seco et al. [16] were the first one computed IC with ontology hierarchical structure. They discovered that the IC of concept subsumed more child nodes was fewer and the IC of each of its leaf nodes was larger in a classification tree. The calculating method of IC value is as follows [17]:

$$IC(c) = 1 - \frac{Log(|hypo(c)|+1)}{Log(max\_nodes)} \quad (4)$$

Where $hypo(c)$ is the count of child nodes of node $c$, $max\_nodes$ represents the maximum number of the concepts in the classification tree. It can be seen from (4) that the IC was only related to the intrinsic hierarchical structure, and the IC value of node $c$ could be computed by the number of hyponym of node $c$.

Later, David Sanchez et al. [18] proposed a new model, which adopted *subsumers* of leaf node to calculate the value of IC. The equation was as follows:

$$IC_{David}(c) = -log\left(\frac{commonness(c)}{commonness(root)}\right) \quad (5)$$

Where the function *commonness(c)* equals $\sum commonness(n)$, and *commonness(n)* equals *1/subsumers(n)*. Wherein $n$ is a leaf node and one of hyponym of node $c$, *subsumers(n)* returns the number of nodes from the *root* to node $n$ along the path of taxonomy.

From above we concluded that the method based on ontology intrinsic structure only relied on the hierarchical structure without any external information, so it was more accurate than statistical method.

## 2.2 IC-based semantic similarity measures

Typical semantic similarity measures include Resnik's [12], Jiang and Conrath's [13] and Lin's measure [14] etc.

Resnik [12] was the first one computed semantic similarity through the intrinsic structure in ontology, which judged the similarity of a pair of concepts by the amount of sharing information. Therefore he regarded the Most Specific Common Abstraction (MSCA) that subsumes both concepts as the semantic similarities of the two concepts. The model is as follows [12]:

$$sim_R(c_1, c_2) = -lg\ p(lso(c_1, c_2)) = IC(lso(c_1, c_2)) \quad (6)$$

Here, $lso(c_1, c_2)$ stand for the MSCA of $c_1$ and $c_2$ in taxonomy.

Jiang & Conrath [13] computed the semantic distance through the IC sum of two concepts subtracting the IC of their MSCA. The measure is as follows:

$$dist_{JC}(c_1, c_2) = IC(c_1) + IC(c_2) - 2 \times IC(lso(c_1, c_2)) \quad (7)$$

After a linear transformation, the equation (7) could be transformed as follows [16].

$$sim_{J\&c}(c_1, c_2) = 1 - \left( \frac{IC(c_1) + IC(c_2) - 2 \times IC(lso(c_1, c_2))}{2} \right) \quad (8)$$

Lin [14] had a same understanding as Resnik on semantic similarity, and he believed that the similarity of two concepts could be measured by the ratio of common information and total information. The model proposed by him is as follows [14]:

$$sim_{Lin}(c_1, c_2) = \frac{2 \times IC(lso(c_1, c_2))}{IC(c_1) + IC(c_2)} \quad (9)$$

Based on stated above, it is noted that, the IC-based similarity measures, the most critical issue was how to get the IC value of concept exactly and how to introduce IC into the measures.

## 2.3 Path and depth based measures

Except for IC-based measures, there are some other representative semantic similarity measures, including Rada's [18], Wu & Palmer's [19] and Leacock & Chodorow's [20] etc.

Rada et al. [18] stated that the length of the minimum path of two concepts quantified their semantic distance. Namely, the similarity between words can be calculated by the minimum path distance linking their corresponding nodes of "is-a" links of ontology. A simple measure to calculate their semantic distance defined by [18] is:

$$dis_{rad}(c_1, c_2) = \min_{\forall i} |path_i(c_1, c_2)| \quad (10)$$

Wu & Palmer's measure [19] is a typical method based on the shortest path. Their method adopted depth and length to compute the similarity. The corresponding calculating equation is as follows [19]:

$$sim_{W\&P}(c_1, c_2) = \frac{2 \times depth(lso(c_1, c_2))}{len(c_1, c_2) + 2 \times depth(lso(c_1, c_2))} \quad (11)$$

Where the function $depth(c_i)$ is the depth of $c_i$. $len(c_1, c_2)$ stand for the shortest path distance between $c_1$ and $c_2$. $lso(c_1,c_2)$ represent the MSCA of $c_1$ and $c_2$.

Later, Leacock & Chodorow [20] proposed a non-linear calculating model, which included two parameters. One is the number of nodes between two concepts (including itself), and the other is the maximum depth of the classification tree. The calculating equation is as follows [20]:

$$sim_{L\&C}(c_1,c_2) = -\log \frac{len(c_1,c_2)}{2 \times max_{c \in WordNet} depth(c)} \quad (12)$$

For a fixed classification tree, we can see in the equation (12), if the path distance between two concepts was further, the semantic similarity was smaller.

### 3 A new hybrid measure of concept semantic similarity

As stated section 2, IC-based semantic similarity measures [12–15] include two parts: IC computing method and IC-based similarity measures [10]. Computing IC is the key of similarity computation of IC-based, so in this section we focus on two parts: improving IC model and proposing a new semantic similarity measure.

### 3.1 Rewriting the IC model by introducing information theory

As Rada et al. [18] stated, the length of the minimum path of two concepts quantified their semantic distance.

$$dis_{rad}(c_1,c_2) = |min\_path(c_1,c_2)| \quad (13)$$

According to information theory of the paper [9], the deeper the concept is, the greater the information content is. As stated in [21], differential information of a content comparing to another could be quantified by the IC of the concept alone subtracting the public parts of two information content. The calculating equation is as follows [9]:

$$\begin{aligned}|min\_path(c_1,c_2)| &= length(c_1,c_2) \\ &\cong (IC(c_1)-IC(c_2))+(IC(c_2)-IC(c_1)) \\ &= (IC(c_1)-IC(lso(c_1,c_2)))+(IC(c_2)-IC(lso(c_1,c_2))) \\ &= IC(c_1)+IC(c_2)-2 \times IC(lso(c_1,c_2)) \end{aligned} \quad (14)$$

Where the function $IC(lso(c_1,c_2))$ are the IC of the MSCA of $c_1$ and $c_2$. As stated above, a conceptual relative depth is the minimum distance between the concept and root, and root is the MSCA between root node and any node, namely $IC(lso(c, root))=IC(root)$. As the root includes any concept, namely $IC(root)=0$, the depth of concept $c$ can be approximated as follows [22]:

$$\begin{aligned}depth(c) &= min\_path(c,root) \\ &\cong IC(c)+IC(root)-2 \times IC(lso(c,root)) \\ &= IC(c)-IC(root) = IC(c) \end{aligned} \quad (15)$$

In the same way, the equation (5) could be rewritten as following:

$$IC_{new}(c) = -\log\left(\frac{commonness(c)}{commonness(root)}\right) = \log(subsumers(c)) \quad (16)$$

Where the function *commonness(c)* equals $\sum commonness(n)$, and *commonness(n)* equals *1/subsumers(n)*, in which $n$ is a leaf node and one of hyponym of $c$. The function *subsumers(c)* equals ∑*subsumers(n)*, and *subsumers(n)* returns the number of node from the *root* to node *n* along the path of taxonomy, namely the number of direct hypernym of $c_i$ (including $c_i$ itself). Because the function *commonness(root)* equals *1/subsumers(root)* and *subsumers(root)* equals 1, therefore the function *commonness (root)* equals 1.

The model of equation (16) introduced the information theory and included intrinsic parameters the number of child-nodes of concept *c* and the depth of concept *c* in taxonomy, so this IC method belonged to a hybrid computing IC method. In addition, this IC model does not be interfered by external factors, therefore in theory it can achieve better stability.

### 3.2 A comprehensive semantic similarity approach between words

Before propose our similarity approach, we define four definitions of semantic similarity as follows.

**Definition 1. Hypernyms** $hyp(c)=\{c_i \in V, c \in V | c < c_i\} \cup \{c\}$ . In this equation, $c_i$ stands for the nodes along the path from the *root* to node *c* in the classification tree, *V* is the set of concepts of the classification tree.

**Definition 2. Directhypernyms** $dhyp(c)=\{c_i \in V, c_i \rightarrow c\}$, $c_i$ is the direct hypernym of *c* (including *c* itself) , *V* is a set of concepts of the classification tree.

**Definition 3. Lowest Common Hypernym** $lch(c_1, c_2)$ the most specific common abstraction that subsumes both concepts $c_1$ and $c_2$.

**Definition 4. Max Depth** $c_{max\_depth}$ represents the depth of the deepest node in the classification tree.

Based on the equation (16) and Leacock & Chodorow similarity measure (12), we propose a comprehensive semantic similarity approach which took into account the information theory and the taxonomy structure. The new semantic similarity approach is as follows:

$$sim_{new}(c_1,c_2) = log \frac{2 \times log(dhyp(c_{max\_depth}))}{log(dhyp(c_1)) + log(dhyp(c_2)) - 2 \times log(dhyp(lch(c_1,c_2)))} \quad (17)$$

In theory, our measure belonged to the comprehensive measure because our approach integrates characters of IC-based measure and feature-based measure. Our approach owns three advantages. Firstly, this method does not interfere by external factors because it is based on ontology intrinsic structure. Secondly, this approach reduced the count of parameter because it only included one parameter (*dhyp* and *lch*). Thirdly, this approach simplified the difficulty of operation by converting the minimum distance between $c_1$ and $c_2$ to direct hypernym of the most specific common abstraction of $c_1$ and $c_2$. In part 4, we will design an experiment to compare the semantic similarity measures in a bench mark dataset.

### 4 Semantic similarity measures evaluation

As Adhikari *et al.* stated [3], the first step of finding semantic similarity is designing a good IC computing model and the second step is using the IC computing model in an efficient similarity measure. In order to evaluate the proposed IC model and our new similarity measure, we will compare our measure and four typical semantic similarity measures mentioned in section 2.

### 4.1 Data source and concept selection

In this paper, we compute the semantic similarity degree between words by the ontology of

WordNet 3.0 version, and make an experiment on Rubenstein & Goodenough (R&G) benchmark dataset [23].

WordNet 3.0 [4] was organized in a taxonomical way and included more than ten thousands of English concepts. In WordNet 3.0 each word was described by a set of concepts that express the possible meanings of the concerned word. The taxonomy "is-a" was mainly semantic relations and was used to compute the semantic similarity degree between words, which were more important in semantic computing. In WordNet 3.0, because noun reached the 75 percent, so our measure used the nominal "is a" taxonomies of WordNet in this paper.

R&G dataset included 63 word pairs, which were judged by 51 professional people. We chose 30 word-pairs, and the range of similarity was from irrelevant to identity. According to the similarity degree between words, artificial scoring range was in [0.0-4.0].

We computed the semantic similarity in WordNet 3.0 by an accepted website, which includes Jiang & Conrath's, Resnik's, Lin's, Wu & Palmer's, Leacock & Chodorow's measures and so on [24]. Considering the situation that each word corresponds to a number of concepts in WordNet and R&G dataset only include words, so we need to transform seeking concept into seeking word. We assumed word $w_1$ owns $m$ concepts and $w_2$ owns $n$ concepts. When calculating the similarities of $w_1$ and $w_2$, we could get $m \times n$ similarity values. Wherein, we adopted the largest value of a word as the concept semantic similarity. We gave a specific model for seeking conceptual similarities as follows:

$$sim(w_1, w_2) = \max_{(i,j)}[sim(c_{1i}, c_{2j})] \qquad (16)$$

Here, $c_{1i}$ is the concept of $w_1$, and $c_{2j}$ is the concept of $w_2$.

**4.2 Evaluation metrics**

Seeking the correlation coefficient of similarity measure and artificial data is an important bench mark for evaluating similarity measure. We evaluated our approach by the equation (17) (two-sided 0.05 level Pearson correlation measurement) [6, 25]:

$$r_{xy} = \frac{\sum_{i=1}^{n}(x_i - \bar{x})(y_i - \bar{y})}{\sqrt{\sum_{i=1}^{n}(x_i - \bar{x})^2(y_i - \bar{y})^2}} \qquad (17)$$

Here, X represents similarity value computed by a similarity measure in R&G and Y represents similarity value derived from artificial data in R&G, and Y is used in the benchmark data to evaluate similarity measures. X equals $(x_1, x_2, ...x_n)$ and Y equals $(y_1, y_2, ...y_n)$. $(x_i - \bar{x})$ is the difference between $x_i$ and mean of $x_i$, and $x_i$ represents each term of set X. Similarly, $(y_i - \bar{y})$ is the difference between $y_i$ and mean of $y_i$, and $y_i$ represents each term of set Y. The correlation coefficient $r_{xy}$ is in [1, -1].

**5 Experimental and evaluation results**

We design an experiment to test similarity scores of 30 term pairs on Wu & Palmer's, Jiang & Conrath's, Leacock & Chodorow's, artificial data, Lin's and our measure.

**Table 1 Comparing similarity scores for different similarity measures in set of 30 term pairs of R&G dataset**

| Word-pair | Artificial | Wu & | Jiang & | Lin's | Leacock & | Sim(new) |
| --- | --- | --- | --- | --- | --- | --- |

|  | data | Palmer's | Conrath's |  | Chodorow's |  |
|---|---|---|---|---|---|---|
| autograph-shore | 0.0600 | 0.3077 | 0.0000 | 0.0000 | 1.3863 | 0.2188 |
| noon-string | 0.0800 | 0.3529 | 0.0653 | 0.0923 | 1.2040 | 0.3815 |
| glass-magician | 0.1100 | 0.5333 | 0.0604 | 0.1421 | 1.6094 | 0.3852 |
| automobile-wizard | 0.1100 | 0.4545 | 0.0738 | 0.1682 | 1.1239 | 0.4930 |
| mound-stove | 0.1400 | 0.6667 | 0.0681 | 0.3143 | 1.7430 | 0.5976 |
| coast-forest | 0.4200 | 0.6154 | 0.0628 | 0.1181 | 1.8971 | 0.6056 |
| boy-rooster | 0.4400 | 0.5600 | 0.0727 | 0.2094 | 1.2040 | 0.6849 |
| cushion-jewel | 0.4500 | 0.6667 | 0.0694 | 0.2572 | 1.7430 | 0.7610 |
| coast-hill | 0.8700 | 0.7143 | 0.2187 | 0.7286 | 2.0794 | 0.7622 |
| boy-sage | 0.9600 | 0.6667 | 0.0680 | 0.2057 | 1.8971 | 0.9472 |
| mound-shore | 0.9700 | 0.7143 | 0.1672 | 0.6724 | 2.0794 | 0.8045 |
| automobile-cushion | 0.9700 | 0.6364 | 0.0894 | 0.3812 | 1.5404 | 0.7208 |
| crane-rooster | 1.4100 | 0.7586 | 0.0000 | 0.0000 | 1.6094 | 0.9923 |
| hill-woodland | 1.4800 | 0.6154 | 0.0592 | 0.1218 | 1.8971 | 0.6056 |
| brother-lad | 1.6600 | 0.7143 | 0.0830 | 0.2400 | 2.0794 | 1.0136 |
| crane-implement | 1.6800 | 0.7778 | 0.0784 | 0.3327 | 2.0794 | 0.9729 |
| magician-oracle | 1.8200 | 0.6250 | 0.0588 | 0.1828 | 1.7430 | 0.8980 |
| sage-wizard | 2.4600 | 0.1667 | 0.0580 | 0.1809 | 1.8971 | 0.9472 |
| oracle-sage | 2.6100 | 0.7059 | 0.1083 | 0.5885 | 1.8971 | 1.0076 |
| brother-monk | 2.8200 | 0.9565 | 0.0689 | 0.2079 | 2.9957 | 0.8367 |
| implement-tool | 2.9500 | 0.9412 | 0.8484 | 0.9146 | 2.9957 | 1.5718 |
| bird-crane | 2.9700 | 0.8800 | 0.0000 | 0.0000 | 2.3026 | 1.3402 |
| bird-cock | 3.0500 | 0.9565 | 0.2681 | 0.7881 | 2.9957 | 1.7568 |
| hill-mound | 3.2900 | 1.0000 | 0.4931 | 1.0000 | 3.6889 | 1.1924 |
| cord-string | 3.4100 | 0.9412 | 0.6553 | 0.9188 | 2.9957 | 1.0576 |
| midday-noon | 3.4200 | 1.0000 | 3.5685 | 1.0000 | 3.6889 | 1.4007 |
| glass-tumbler | 3.4500 | 0.5882 | 0.0626 | 0.1858 | 1.6094 | 1.1306 |
| serf-slave | 3.4600 | 0.8000 | 0.0000 | 0.0000 | 2.3026 | 0.9776 |
| cemetery-graveyard | 3.8800 | 1.0000 | 1.0000 | 1.0000 | 3.6889 | 1.3395 |
| magician-wizard | 3.5000 | 1.0000 | 0.0640 | 1.0000 | 3.6889 | 1.1110 |
| **range** | **3.8200** | **0.6923** | **1.0000** | **1.0000** | **2.5650** | **1.5380** |

In this study, in the same settings we have evaluated representative measurements generated by machine, and compared them against human ratings performed on semantic similarities between words.

We evaluated our approach by the equation (17) (two-sided 0.05 level Pearson correlation measurement). Testing results of Poisson correlation coefficient showed as follows:

**Table 2 Comparison of correlation coefficient between existed measures and our measure**

| Method | Pearson correlation coefficient |
|---|---|
| Artificial Data | 1 |
| Wu & Palmer Measure | 0.678 |
| Lin Measure | 0.543 |
| Jiang & Conrath Measure | 0.389 |
| Leacock & Chodorow Measure | 0.792 |
| Our Measure | 0.823 |

In Table 2, Poisson correlation coefficient represents the correlation score between each machine generated measure and artificial data.

## 6 Discussions

There were four aspects of our work have to be addressed. Firstly, we rewrote the IC model which we took into account the taxonomy structure and information theory, and converted the old computing IC method (based ontology intrinsic structure) to a new hybrid computing IC method. Based on the new IC model, we proposed a novel semantic similarity approach, which computed the similarities between words based on the comprehensive factors instead of relying on the path and depth. In this similarity approach, we adopted the parameters *lch* and *dhyp*, which validated the estimation of semantic similarity degree between words.

Secondly, the results of Table 1 show that our quantification approach owned better performance than other methods in set of 30 term pairs of R&G dataset. This performance is very desirable because WordNet is a common ontology and the intrinsic IC models of ontology-based own better independence than domain corpora. In fact, the intrinsic IC models are efficient and easily applicable to different domains because the IC models of corpora-based are hampered by corpora used [1].

Thirdly, the parameter *range* is an important benchmark for the dispersion degree of measures. In Table 1, the last row showed that the *range* of our measure reachs 1.5380, and this meant that the dispersion degree of our measure is better than Wu & Palmer's, Jiang & Conrath's, Lin's measures. The dispersion degree of our measure is lower than Leacock & Chodorow's measure (W&P measure *range*=0.6923, L&C measure *range*=2.5652, Lin measure *range*=1.0000, J&C measure *r*=1.0000). This is due to the smallest similarity of word reached the 1.2040 in Leacock & Chodorow's measure, but the smallest similarity of word equaled 0.2188 in proposed measure (Leacock & Chodorow measure "*noon-string*" =1.2040, our measure "*autograph-shore*" =0.2188).

The last point, based on evaluation metrics equation (17), if correlation score was bigger, testing measure and artificial data are closer related. In Table 2, the measures based on path and depth owned good correlations than pure IC-based measures (W&P measure *r*=0.678, L&C measure *r*=0.792; Lin measure r=0.543, J&C measure *r*=0.389). When we introduced our hybrid IC model, the correlation coefficient of our measure reached 0.823. This means the fitting degree of our measure is better than others.

## 7 conclusion and future works

In this paper, our works included three aspects. Firstly, after analyzing representative IC models and typical semantic similarity measures, we proposed an improved computing IC model. Our model has been considered information theory and intrinsic ontology structure factors (hypernym, hyponym, depth, node number), which owned significant weight on computing accurately information content of concepts. Secondly, based on the improved IC model we put forward a new comprehensive measure for estimating concept semantic similarity. Thirdly, we tested our approach in set of 30 term pairs of R&G benchmark dataset and compared the correlation coefficient between existed measures, our measure and artificial data. The results showed our measure was effective. In future, we will improve the proposed approach by considering the more spatial structure of ontology and proof-test this approach in some widely datasets.

**Acknowledgments**

The work in this paper was supported by Chinese National Natural Science Foundation (Grant No. 61562072).